%% file: _main.tex
\documentclass[conference]{IEEEtran}
\IEEEoverridecommandlockouts
\usepackage{cite}
\usepackage{amsmath,amssymb,amsfonts}
\usepackage{algorithmic}
\usepackage{graphicx}
\usepackage{textcomp}
\usepackage{xcolor}
\def\BibTeX{{\rm B\kern-.05em{\sc i\kern-.025em b}\kern-.08em
    T\kern-.1667em\lower.7ex\hbox{E}\kern-.125emX}}
\begin{document}

\title{Validation of a Hospital Digital Twin with Machine Learning}

\makeatletter
\newcommand{\linebreakand}{%
  \end{@IEEEauthorhalign}
  \hfill\mbox{}\par
  \mbox{}\hfill\begin{@IEEEauthorhalign}
}
\makeatother

\author{\IEEEauthorblockN{Muhammad Aurangzeb Ahmad}
\IEEEauthorblockA{\textit{Department of Computer Science} \\
\textit{University of Washington Bothell}\\
Bothell, WA\\
maahmad@uw.edu}
\and
\IEEEauthorblockN{Vijay Chickarmane}
\IEEEauthorblockA{\textit{ACE AI (of KPInsight)} \\
\textit{Kaiser Permanente}\\
Oakland, USA \\
vijay.s.chickarmane@kp.org}
\and
\IEEEauthorblockN{Farinaz Sabz Ali Pour}
\IEEEauthorblockA{\textit{ACE AI (of KPInsight) } \\
\textit{Kaiser Permanente}\\
Oakland, USA \\
farinaz.x.sabzalipour@kp.org}
\linebreakand 
\IEEEauthorblockN{Nima Shariari}
\IEEEauthorblockA{\textit{ACE AI (of KPInsight) } \\
\textit{Kaiser Permanente}\\
Oakland, USA \\
nima.x.shahriari@kp.org}
\and
\IEEEauthorblockN{Taposh Dutta Roy}
\IEEEauthorblockA{\textit{ACE AI (of KPInsight)} \\
\textit{Kaiser Permanente}\\
Oakland, USA \\
taposh.d.roy@kp.org}
}

\maketitle

\begin{abstract}
Recently there has been a surge of interest in developing Digital Twins of process flows in healthcare to better understand bottlenecks and areas of improvement. A key challenge is in the validation process. We describe a work in progress for a digital twin using an agent based simulation model for determining bed turnaround time for patients in hospitals. We employ a strategy using machine learning for validating the model and implementing sensitivity analysis. 
\end{abstract}


\input{01_intro}
\input{02_sm_ml}
\input{03_related}
\input{04_experiments}
\input{06_conclusion}
\bibliography{ref}

\section*{Acknowledgment}
We would like to acknowledge Dr. Stephen Parodi, Dr. Chethna Vijay, Mitchell Winnik , Dr. Yu-te Lee,  Tanya Scott, Vivian Tan, Sabrina Dahlgren, Wendy Lin, Mike Page, Sean Schuller, Nitin Roy, and Ilker Yaramis for their contributions to this work.

\end{document}

%% file: 01_intro.tex
\section{Introduction}

Digital twins are virtual copies or simulations of systems. These systems can be used throughout the life-cycle of the physical system being digitized from their inception to decommissioning. Digital twins are often used in to optimize operations, reduce costs, and improve efficiency. They can be used to test and optimize the design of a system before it is built, to monitor and diagnose problems with the system while it is in operation, and to predict and prevent failures. Digital twins are increasingly being applied in various fields such as manufacturing, healthcare, public health and governance and meteorology \cite{tao2019make}\cite{jiang2021industrial}. 

Validation of digital twins and simulation models in general pose a number of challenges. While simulations can be validated on retrospective longitudinal data, alternate scenarios by definition are not present in the ground truth. In this paper we investigate how machine learning could be used for validation of simulations in the context of large hospital systems. Simulations are often used in scenarios where there is some knowledge about a phenomenon but not enough information regarding how the outputs of a system would change for a given sub-space of inputs or perturbations since these have not been observed before. Consider how patterns of resource usage changed during COVID-19 which rendered most prediction models ineffective.

\begin{figure*}[htbp]
\centerline{\includegraphics[width=17cm]{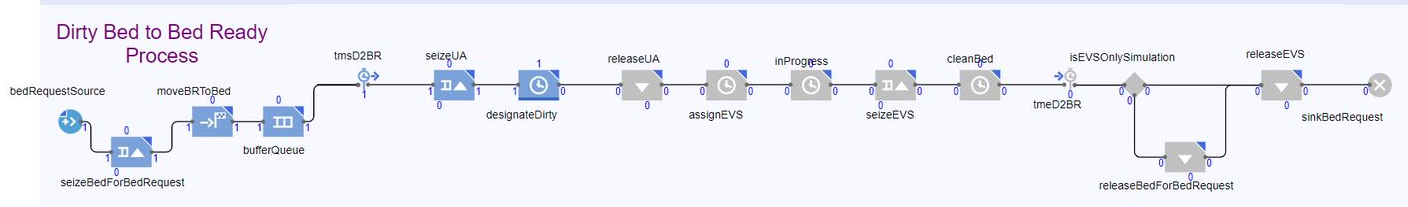}}
\caption{Dirty Bed to Bed Ready Process Simulation Model}
\label{fig:sim}
\end{figure*}

The goal of employing machine learning models in the context of simulations is two-fold. First, to create a model that can predict the outcomes with some level of fidelity and can be used as a benchmark to compare simulated outcomes based on the same parameters that the ML model used. Second, to use machine learning towards sensitivity analysis of the simulation which is normally computationally intensive \cite{kleijnen2005overview}. This is achieved in 2 steps. (1) The simulation model is used to generate synthetic data that can be used to train a ML model. (2) Post-hoc explanation models like SHAP \cite{lundberg2017consistent} can be employed to then determine contributions from individual parameters/features.

This winter (Nov 2022- Feb 2023), Covid-19, RSV, FLU and other ILI (Influenza Like Illness) have pushed our hospital systems to the brink.  Studying the flow of patients from their arrival into the in-patient settings to bed being available to next patient, exposes the various bottlenecks in the process and opportunities for improvement. In the past, hospitals performed process improvement (PI) efforts to address this. However, these efforts take a long time and are generally limited to a small subset of questions that are considered. Further, they cannot be queried to enable a what-if type scenario. Building a digital twin that models from patient arrival to bed turn around time enables a saleable, continuously running and long term solution. 

A detailed throughput flow displayed in Figure \ref{fig:sim} follows patients as they get treated and finally discharged. Subsequently, the dirty beds they occupied are cleaned and recycled back to the various units in the Hospital. The outcome of interest is the bed turnaround time (BTT), which indicates how efficiently  beds can be made available to the next waiting patient.  Understanding the factors that influence BTT  is critical to minimizing any bottlenecks in the flow of clean beds and hence ultimately serving more patients quickly. The goal of this study is to develop a framework that can be used to validate a digital twin model of discharge to bed ready in hospitals. The model will later be adapted according to feedback from operational leadership for individual hospitals. This framework can be used for validation as the simulation evolves. The main contribution of this paper are as follows:
\begin{itemize}
    \item Introduce a digital twin for the discharge to bed ready process.
    \item Develop a framework for validating the digital twin utilizing both machine learning and simulation modeling.
\end{itemize}

%% file: 02_sm_ml.tex
\section{Simulation Models, Digital Twins and Machine Learning}
 A simulation model is a replica of a real-world system on the computer and can be used to evaluate ‘what-if?’ scenarios before actually implementing changes in the real system. For example, a simulation model of a hospital's radiology department could be used to better understand the impact that a new Magnetic Resonance Imaging scanner might have on the hospital's quality of service \cite{liu2018role}. The difference between a digital twin and a simulation is scale, although both fall under the rubric of simulation models. Simulations are meant to accurately represent the phenomenon that they are modeling, this is referred to as validation. verification on the other hand corresponds to establishing that the simulation is correctly implemented. Verification answers the question "Have we built the model right?" whereas validation answers the question "Have we built the right model?” \cite{cook2005perform}

One straightforward way to validate simulations is to compare the outputs of the simulation with ground truth i.e., historical data. For new scenarios, which may not exist in the data, we would require another method to validate the output of the simulation. Here AI/ML models that are trained on historical data can be used to provide a prediction for the new scenarios that can then be used to compare to the simulated outcomes.

In machine learning, surrogate models are often used to either create simplified or interpretable models \cite{lundberg2017consistent}. Machine Learning model that is built off of synthetic data can be conceptualized as a surrogate model. Sensitivity analysis \cite{cook2005perform} is an important part of validation. Understanding which parameters of a model can lead to large variations in the outcomes allows the model to be tested against intuition and from experts who know the ground truth of the phenomena we are trying to model. Even in scenarios where sensitivity analysis may not capture non-linear interactions between the various combinations of inputs and outputs, it can still help the end user understand the  relative importance of inputs to the model \cite{lundberg2017consistent}.

Interpretability of predictive models is important factor in adoption of such models in healthcare since many end users e.g., physicians, nurses etc. are interested in knowing what are the driving factors behind predictions \cite{ahmad2018interpretable}. Thus, sensitivity analysis of complex simulation models is computationally hard using traditional methods \cite{tjoa2022quantifying}. We propose that we use model attribution methods widely used in machine learning like SHAP \cite{lundberg2017consistent} for sensitivity analysis for variable importance attribution. Training machine learning models with synthetic data may help the machine learning model learns about the internal causal structure of the phenomenon of interest \cite{tjoa2022quantifying}, as opposed to just using only the historical data.

%% file: 03_related.tex
\section{Related Work}
There is a large body of work on simulation modeling in various domains. It is not possible to cover the literature comprehensively. We refer the reader to \cite{chung2003simulation} and \cite{rossetti2015simulation} for an overview of simulation modeling. One of the earliest works on simulation of hospital systems was published in 1965 by Fetter and Thompson \cite{fetter1965simulation} who first described the problem of using simulations to understand processes in hospital systems. There are multiple surveys of simulation modeling in healthcare:  Klein et. al. \cite{klein1993simulation}, Mielczarek et al \cite{mielczarek2012application}, Arisha et al \cite{arisha2016modeling}, and Vazquez et al \cite{vazquez2021discrete}. While there are several simulation modeling paradigms, much of the work in simulation modeling in healthcare has been done in discrete event simulation.  

Applications of simulation modeling in healthcare include patient admission models \cite{mielczarek2012application}, patient flow in emergency rooms \cite{brenner2010modeling}, bed utilization in hospitals and clinics \cite{moengin2014discrete}, modeling chronic conditions \cite{mielczarek2012application}, allocation of human resources in hospital systems \cite{tao2019make}. There is some previous work in combining simulation models with machine learning approaches in healthcare. Elbattah et al \cite{elbattah2016coupling} describe approaches for coupling machine learning with simulation modeling for elderly discharge planning. Olave-Rojas et al describe a hybrid model for pre-hospital Emergency Medical Services for combining machine learning and simulation models. Mivsic et al  \cite{mivsic2021simulation} employed simulation methods for evaluation of machine learning systems for hospital readmission systems.

\begin{table}[]
\centering
\begin{tabular}{|l|r|r|r|r|}
\hline
\textbf{Facility} & \textbf{$MAE_{ML}$} &  \textbf{$MAE_{Sim}$} & \textbf{$Sim_{1SD}$} & \textbf{$Sim_{2SD}$}  \\ \hline
All    & 31.99 &  20.87 & 0.97 & 0.99\\ \hline  
Facility 1 & 20.11 & 18.47 & 0.99 & 1.00\\ \hline 
Facility 2 & 23.18 & 20.47 & 0.97 & 0.99\\ \hline 
Facility 3 & 24.08 & 37.77 & 0.97 & 0.98\\ \hline 
Facility 4 & 49.94 & 17.47 & 0.94 & 0.99\\ \hline 
Facility 5 & 51.87 & 21.09 & 0.99 & 0.99\\ \hline 
Facility 6 & 22.49 & 9.88 & 0.96 & 1.00\\ \hline
\end{tabular}
\caption{Overall Model Performance across facilities}
\label{tab:results}
\end{table}

%% file: 04_experiments.tex
\section{Experiments and Results}
Figure \ref{fig:sim} shows the \textit{Discharge to Bed Ready Model} which simulates the process by which dirty beds get cleaned and are readied for the next patient. The process flow is as follows:
 After the patient is discharged the dirty bed that was being occupied is designated as dirty by a Unit assistant(UA) resource. An Environmental services(EVS) resource is then assigned to clean the bed. The target variable, Bed Turnaround Time (BTT), corresponds to how much time does it take for a bed, previously occupied by a patient, to be cleaned. 

\subsection{Data}
The dataset spans from April 11, 2021 to March 30, 2022. The data is available for 6 different hospital facilities and prediction is also done at the facility level. The features used in the ML/simulation models are daily averages of the following:
\begin{itemize}
    \item Number of discharges during the Morning/Evening/Night shifts ("day", "eve", "night)
    \item Number of Unit Assistant (UA) resources during the Morning/Evening/Night shifts ("day ua", "eve ua", "night ua")
    \item Number of Environmental resources (EVS) resources during the Morning/Evening/Night shifts ("day ",evs "eve evs", "night evs")
    \item Time steps for the cleaning process (4 steps)
    \begin{itemize}
    \item time for dirty bed to be assigned ("Avg Dirty Wait Duration")
    \item time for EVS resource to be assigned for cleaning ("Avg Assigned Wait Duration")
    \item  time for bed to be cleaned by an EVS resource ("Avg Clean Wait Duration")
    \item  time for clean bed to be recycled back into  "Avg In Progress Wait Duration"
    \end{itemize} 
\end{itemize}

 

\subsection{Model Setup}
\subsubsection{Simulation Model}
The simulation model is implemented in AnyLogic \cite{borshchev2014multi} which is a widely used simulation software. A schema of the simulation is given in figure \ref{fig:sim} which shows that the simulation is modeling two interrelated processes: patients arrival to wheel out process and dirty bed to bed ready process. The simulations are stochastic in nature and for any set of inputs the simulation is run multiple times so that the output is not just a single output but rather set of outputs. For comparison with the ground truth we use the mean of the output as well as the standard deviation.

\begin{figure}[htbp]
\centerline{\includegraphics[width=8cm]{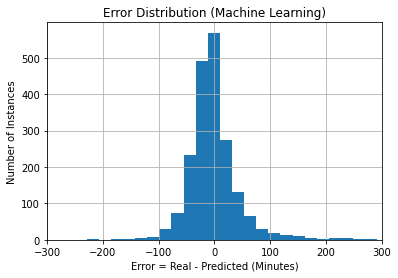}}
\caption{Error Distribution of Machine Learning Model}
\label{fig:ml_error}
\end{figure}

\begin{figure}[htbp]
\centerline{\includegraphics[width=8cm]{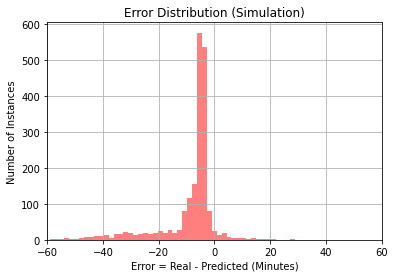}}
\caption{Error Distribution of Simulation Model}
\label{fig:sim_error}
\end{figure}

The Hospital Throughput Simulation which is the focal point of the digital twin system, mimics the patient flow process as follows: (i) Patients are admitted from the Emergency Department (ED), Direct Admissions (DA) and Operations Room (OR) into the ICU, Medical Surgery Department or the Medical Telemetry Department. (ii) After treatment, patients either transition  among units or get discharged after a discharge order is given. (iii) The dirty beds are cleaned and recycled back into the units for newly arriving patients.

The agent based  model simulates each dirty bed coming in randomly in time during each of the shifts sampled from an empirical distribution based on Hospital data. The data also includes the average number of discharges per shift per day. For each bed that is cleaned there are 4 time steps which correspond to the steps in the cleaning process during which the UA and EVS resources are utilized. The time spent in each stage of the cleaning process is also sampled from empirical data. The simulation is typically implemented for 100 days which allows the collection of enough observations for analysis. The target variable is the time taken between when the dirty bed first gets assigned to be cleaned till it is finally cleaned which is the bed turn around time (BTT). A bottleneck can occur in the process flow when several discharges occur within a short time, since there are limited resources and each step takes time. This leads to a delay in the total time it takes for a bed to be cleaned (after averaging over the simulation time period).

\subsubsection{Machine Learning}
To give confidence in the simulation, one would require a comparison of the simulated BTT average with another benchmark. We trained an ML model on the same data that was an input to the simulation and regressed against the actual BTT observed. This ML model was then used to generate predictions for new scenarios (new values of the input parameters) and can serve as a benchmark for comparison with the simulated BTT.  We used a number of regression prediction models but obtained the best results from Gradient Boosting Regressor. 
 

\subsection{Results}
\subsubsection{Comparison with Historical Data}
We used the historical empirical data as inputs to the simulation model. On average, one simulation take about one minutes and thus it took several days to run the simulations for the whole dataset. BTT of the simulation is then compared to the historical BTT. In addition to MAE, we also look at how often does the Actual BTT falls within 1-2 $\sigma $standard deviation (SD) of the simulated BTT. Table \ref{tab:results} gives a summary of the prediction results.  \textit{Sim 1D} and \textit{Sim 2D} correspond to whether the prediction was within two and three standard deviations from the prediction. The results suggest that the simulation is able to cover $> 94 \%$ 
of the actual results.

\subsubsection{Comparison with ML Predictions}
In Table \ref{tab:results} we see that the errors for the ML model are in the same range as the simulation which allows comparison of simulated outcomes with ML predictions for new scenarios. The reason why in some facilities the ML results are worse off than the simulated ones can be attributed to both insufficient data as well as the quality of the data. This is also the reason that the simulation MAE is better that the ML MAE overall. A few facilities which have low quality data bias the model in making poorer predictions. In the future we hope to speed up simulations which will not limit the number of instances that we will be able to generate, thereby giving us more data. Data quality will also improve as more facilities embrace digital reporting. In Figure \ref{fig:ml_error} and Figure \ref{fig:sim_error} we plot the error distribution, where an error defined as $error=actual - predicted$, for the ML model and for the simulation. As compared to the ML error simulation error is skewed suggesting that the simulation tends to over-predict. An analysis of the outliers (error $<$ -60) showed that the inputs into the simulation have a smaller number of EVS resources as compared to the "normal" cases. Incoming dirty beds have to wait in the queue before each EVS resource finishes cleaning the current bed that the resource has been assigned. If there are fewer EVS resources then this process would take more time as it cannot run in parallel as opposed to when there are additional resources.


\begin{figure}[htbp]
\centerline{\includegraphics[width=9cm]{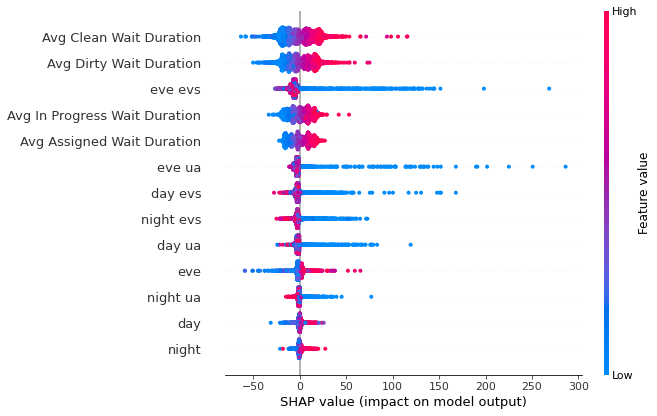}}
\caption{Model Explanation for the Simulation Model}
\label{fig:shap_sm}
\end{figure}

\subsubsection{Sensitivity Analysis through ML models}
Sensitivity analysis, which model explanations are a subset of, enables ascription of change in parameters to the outcomes\cite{niida2019sensitivity}. This allows a domain expert to determine which factors have the largest impact on the outcomes. Afterwards the domain expert, in our case the hospital operational leadership team, has the choice of the levers that can be used to optimize the outcomes. However, in large simulations there are multiple parameters which make sensitivity analysis a computational challenge\cite{saltelli2008global} \cite{sobol2001global}. By using simulated data to train ML models, we can test global sensitivity of the model parameters through attribution analysis such as SHAP \cite{lundberg2017consistent}.

Figure \ref{fig:shap_sm} shows the relative importance of input variables at the global level for the simulation model.  It shows that the time steps for bed cleaning and evening EVS resources are most important. Whereas an increase in the time steps raises the BTT, a decrease in EVS resources leads to the same outcome. Number of evening discharges have a higher impact on BTT than discharges during the day or night. This accords with the intuition around how patient flow in such departments work. One can also observe that the impact of the average clean wait duration on BTT is linear. Adding more EVS resources reduces BTT but its effect saturates beyond a certain limit.

%% file: 06_conclusion.tex
\section{Conclusion and Future Work}
In this paper we described how machine learning can augment simulation modeling for a digital twin system. The specific case we described was the process of discharge to bed ready, in which after a patient is discharges, the dirty bed that was occupied is cleaned and made available to the next patient. We used historical data from six hospital systems to validate the models. We are working closely with the hospital operations leadership team to deploy these models in a real world setting where they will be used for decision making for resource allocation.

In the future we plan to expand the current framework to include a Full Hospital Simulation. We plan to extent this work to a Multi-Outcome regression problem, which will include patient wait times and other metrics which relate to hospital operational efficiency. We also plan to use emulation for expanding the scope of the sensitivity analysis and automating parts of the validation framework to take into account updates in model as well as input data. 

%% file: _main.bbl
\newpage


\begin{thebibliography}{10}

\bibitem{bica2021real}
Ioana Bica, Ahmed~M Alaa, Craig Lambert, and Mihaela Van Der~Schaar.
\newblock From real-world patient data to individualized treatment effects
  using machine learning: current and future methods to address underlying
  challenges.
\newblock {\em Clinical Pharmacology \& Therapeutics}, 109(1):87--100, 2021.

\bibitem{Pearl2000CausalityMR}
Judea Pearl.
\newblock Causality: Models, reasoning and inference.
\newblock 2000.

\bibitem{wagg2020digital}
DJ~Wagg, Keith Worden, RJ~Barthorpe, and Paul Gardner.
\newblock Digital twins: state-of-the-art and future directions for modeling
  and simulation in engineering dynamics applications.
\newblock {\em ASCE-ASME J Risk and Uncert in Engrg Sys Part B Mech Engrg},
  6(3), 2020.

\bibitem{manufact2022DigitalTwin}
Igiri Onaji, Divya Tiwari, Payam Soulatiantork, Boyang Song, and Ashutosh
  Tiwari.
\newblock Digital twin in manufacturing: conceptual framework and case studies.
\newblock {\em International Journal of Computer Integrated Manufacturing},
  35(8):831--858, 2022.

\bibitem{VANDERVALK2022156}
Hendrik {van der Valk}, Gero Strobel, Stephanie Winkelmann, Joachim Hunker, and
  Martin Tomczyk.
\newblock Supply chains in the era of digital twins – a review.
\newblock {\em Procedia Computer Science}, 204:156--163, 2022.
\newblock International Conference on Industry Sciences and Computer Science
  Innovation.

\bibitem{traffic2023DigitalTwin}
Y.~Li and W.~Zhang.
\newblock Traffic flow digital twin generation for highway scenario based on
  radar-camera paired fusion.
\newblock {\em Scientific Reports}, 13(642), 2023.

\bibitem{Chen2020simulation}
Y.~K.~Liu J.~H.~Chen and Y.~H. Kao.
\newblock Digital twins for healthcare: A review.
\newblock {\em Journal of Medical Systems}, 44(12):275, 2020.

\bibitem{tao2019make}
Fei Tao and Qinglin Qi.
\newblock Make more digital twins, 2019.

\bibitem{jiang2021industrial}
Yuchen Jiang, Shen Yin, Kuan Li, Hao Luo, and Okyay Kaynak.
\newblock Industrial applications of digital twins.
\newblock {\em Philosophical Transactions of the Royal Society A},
  379(2207):20200360, 2021.

\bibitem{liu2018role}
Zheng Liu, Norbert Meyendorf, and Nezih Mrad.
\newblock The role of data fusion in predictive maintenance using digital twin.
\newblock In {\em AIP conference proceedings}, volume 1949, page 020023. AIP
  Publishing LLC, 2018.

\bibitem{cook2005perform}
David~A Cook and James~M Skinner.
\newblock How to perform credible verification, validation, and accreditation
  for modeling and simulation.
\newblock {\em The Journal of Defense Software Engineering}, 18(5):20--24,
  2005.

\bibitem{ahmad2018interpretable}
Muhammad~Aurangzeb Ahmad, Carly Eckert, and Ankur Teredesai.
\newblock Interpretable machine learning in healthcare.
\newblock In {\em Proceedings of the 2018 ACM international conference on
  bioinformatics, computational biology, and health informatics}, pages
  559--560, 2018.

\bibitem{lundberg2017consistent}
Scott~M Lundberg and Su-In Lee.
\newblock Consistent feature attribution for tree ensembles.
\newblock {\em arXiv preprint arXiv:1706.06060}, 2017.

\bibitem{chung2003simulation}
Christopher~A Chung.
\newblock {\em Simulation modeling handbook: a practical approach}.
\newblock CRC press, 2003.

\bibitem{rossetti2015simulation}
Manuel~D Rossetti.
\newblock {\em Simulation modeling and Arena}.
\newblock John Wiley \& Sons, 2015.

\bibitem{fetter1965simulation}
Robert~B Fetter and John~D Thompson.
\newblock The simulation of hospital systems.
\newblock {\em Operations research}, 13(5):689--711, 1965.

\bibitem{klein1993simulation}
Robert~W Klein, Robert~S Dittus, Stephen~D Roberts, and James~R Wilson.
\newblock Simulation modeling and health-care decision making.
\newblock {\em Medical decision making}, 13(4):347--354, 1993.

\bibitem{mielczarek2012application}
Bo{\.z}ena Mielczarek and Justyna Uzia{\l}ko-Mydlikowska.
\newblock Application of computer simulation modeling in the health care
  sector: a survey.
\newblock {\em Simulation}, 88(2):197--216, 2012.

\bibitem{arisha2016modeling}
Amr Arisha and Wael Rashwan.
\newblock Modeling of healthcare systems: past, current and future trends.
\newblock In {\em 2016 Winter Simulation Conference (WSC)}, pages 1523--1534.
  IEEE, 2016.

\bibitem{vazquez2021discrete}
Jes{\'u}s~Isaac V{\'a}zquez-Serrano, Rodrigo~E Peimbert-Garc{\'\i}a, and
  Leopoldo~Eduardo C{\'a}rdenas-Barr{\'o}n.
\newblock Discrete-event simulation modeling in healthcare: A comprehensive
  review.
\newblock {\em International Journal of Environmental Research and Public
  Health}, 18(22):12262, 2021.

\bibitem{brenner2010modeling}
Stuart Brenner, Zhen Zeng, Yang Liu, Junwen Wang, Jingshan Li, and Patricia~K
  Howard.
\newblock Modeling and analysis of the emergency department at university of
  kentucky chandler hospital using simulations.
\newblock {\em Journal of emergency nursing}, 36(4):303--310, 2010.

\bibitem{elbattah2016coupling}
Mahmoud Elbattah and Owen Molloy.
\newblock Coupling simulation with machine learning: A hybrid approach for
  elderly discharge planning.
\newblock In {\em Proceedings of the 2016 ACM SIGSIM Conference on Principles
  of Advanced Discrete Simulation}, pages 47--56, 2016.

\bibitem{mivsic2021simulation}
Velibor~V Mi{\v{s}}i{\'c}, Kumar Rajaram, and Eilon Gabel.
\newblock A simulation-based evaluation of machine learning models for clinical
  decision support: application and analysis using hospital readmission.
\newblock {\em NPJ Digital Medicine}, 4(1):1--11, 2021.

\bibitem{kleijnen2005overview}
Jack~PC Kleijnen.
\newblock An overview of the design and analysis of simulation experiments for
  sensitivity analysis.
\newblock {\em European Journal of Operational Research}, 164(2):287--300,
  2005.

\bibitem{mahajan2017hospital}
Aman Mahajan, Salim~D Islam, Michael~J Schwartz, and Maxime Cannesson.
\newblock A hospital is not just a factory, but a complex adaptive
  system—implications for perioperative care.
\newblock {\em Anesthesia \& Analgesia}, 125(1):333--341, 2017.

\bibitem{tjoa2022quantifying}
Erico Tjoa and Guan Cuntai.
\newblock Quantifying explainability of saliency methods in deep neural
  networks with a synthetic dataset.
\newblock {\em IEEE Transactions on Artificial Intelligence}, 2022.

\bibitem{borshchev2014multi}
Andrei Borshchev.
\newblock Multi-method modelling: Anylogic.
\newblock {\em Discrete-Event Simulation and System Dynamics for Management
  Decision Making}, pages 248--279, 2014.

\bibitem{moengin2014discrete}
Parwadi Moengin, Winnie Septiani, and Selvia Herviana.
\newblock A discrete-event simulation methodology to optimize the number of
  beds in hospital.
\newblock In {\em Proceedings of the World Congress on Engineering and Computer
  Science}, volume~2, 2014.

\bibitem{niida2019sensitivity}
Atsushi Niida, Takanori Hasegawa, and Satoru Miyano.
\newblock Sensitivity analysis of agent-based simulation utilizing massively
  parallel computation and interactive data visualization.
\newblock {\em PloS one}, 14(3):e0210678, 2019.


\bibitem{saltelli2008global}
Andrea Saltelli, Marco Ratto, Terry Andres, Francesca Campolongo, Jessica
  Cariboni, Debora Gatelli, Michaela Saisana, and Stefano Tarantola.
\newblock {\em Global sensitivity analysis: the primer}.
\newblock John Wiley \& Sons, 2008.

\bibitem{sobol2001global}
Ilya~M Sobol.
\newblock Global sensitivity indices for nonlinear mathematical models and
  their monte carlo estimates.
\newblock {\em Mathematics and computers in simulation}, 55(1-3):271--280,
  2001.


\end{thebibliography}
